\pdfoutput=1

\documentclass[11pt]{article}
\usepackage[dvipsnames]{xcolor}
\usepackage{ACL2023}
\usepackage{times}
\usepackage[dvipsnames]{xcolor}
\usepackage{latexsym}
\usepackage{booktabs}
\usepackage[T1]{fontenc}
\usepackage[utf8]{inputenc}
\usepackage{microtype}
\usepackage{multirow}
\usepackage{graphicx}
\usepackage{epigraph} 
\usepackage{xspace} 
\usepackage{enumitem}
\usepackage{amsmath}
\usepackage{tikz}
\usepackage{amssymb}
\usepackage{array}
\usepackage{float}
\usepackage{pifont}

\usepackage{fancyhdr}
\usepackage{fixltx2e}

\usepackage{hyperref}

\usetikzlibrary {arrows.meta}
\setenumerate[1]{itemsep=0pt,parsep=0pt,topsep=0pt}
\setitemize[1]{itemsep=0pt,parsep=0pt,topsep=0pt,leftmargin=6pt}
\graphicspath{{./figures}}

\newcommand{\kt}{\textit{Knowing-that}\xspace}
\newcommand{\kh}{\textit{Knowing-how}\xspace}
\newcommand{\dataname}{\texttt{OHO}\xspace}
\newcommand{\rn}{{TARA}\xspace}
\newcommand{\rns}{{TARAs}\xspace}

\makeatletter

\newcommand{\Rmnum}[1]{\expandafter\@slowromancap\romannumeral #1@}
\makeatother

\title{\kh \& \kt: A New Task for Machine Reading Comprehension of User Manuals}

\author{
{Hongru Liang\textsuperscript{1}\quad Jia Liu\textsuperscript{2}\quad Weihong Du\textsuperscript{1}}
\\
{\textbf{Dingnan Jin}\textsuperscript{2}\quad \textbf{Wenqiang Lei}\textsuperscript{1}\thanks{\quad Corresponding author}\quad  \textbf{Zujie Wen}\textsuperscript{2}\quad  \textbf{Jiancheng Lv}\textsuperscript{1}}
 \\
{\textsuperscript{1}College of Computer Science, Sichuan University} \\
\normalsize{\textsuperscript{2}Ant Group, China} \\
\texttt{\{lianghongru, wenqianglei, lvjiancheng\}@scu.edu.cn}\\ 
\texttt{duweihong@stu.scu.edu.cn}\\
\texttt{\{jianiu.lj, dingnan.jdn, zujie.wzj\}@antgroup.com}\\
}

\begin{document}
\maketitle

\begin{abstract}
 The machine reading comprehension (MRC) of user manuals has huge potential in customer service. However, current methods have trouble answering complex questions. Therefore, we introduce the \kh \& \kt task that requires the model to answer factoid-style, procedure-style, and inconsistent questions about user manuals. We resolve this task by jointly representing the s\underline{T}eps and f\underline{A}cts in a g\underline{RA}ph~(\rn), which supports a unified inference of various questions. Towards a systematical benchmarking study, we design a heuristic method to automatically parse user manuals into \rns and build an annotated dataset to test the model's ability in answering real-world questions. Empirical results demonstrate that representing user manuals as \rns is a desired solution for the MRC of user manuals. An in-depth investigation of \rn further sheds light on the issues and broader impacts of 
 future representations of user manuals. We hope our work can move the MRC of user manuals to a more complex and realistic stage. 
 The code and dataset are available in \href{https://github.com/WitcherLeo/Knowing-how-Knowing-that}{https://github.com/WitcherLeo/Knowing-how-Knowing-that}. 
\end{abstract}

\section{Introduction}
\label{sec:intro}
User manuals
are supposed to be helpful in using products, getting involved in promotions, or other goals of interest~\cite{ryle2009concept,chu2017distilling,bombieri2021automatic}. Though well-designed and instructive, they are seldom read by users because ``\textit{Life is too short to read manuals}''~\cite{8154819}. Towards high user satisfaction, professional customer service representatives~(CSRs) are hired to do the reading and answer user questions about the manuals. The machine reading comprehension~(MRC) of user manuals thus has huge potential, as it would not only reduce high labor costs but enable the service ready for customers 24/7. 
\par
\begin{figure}[t]
    \centering
    \includegraphics[width=7cm]{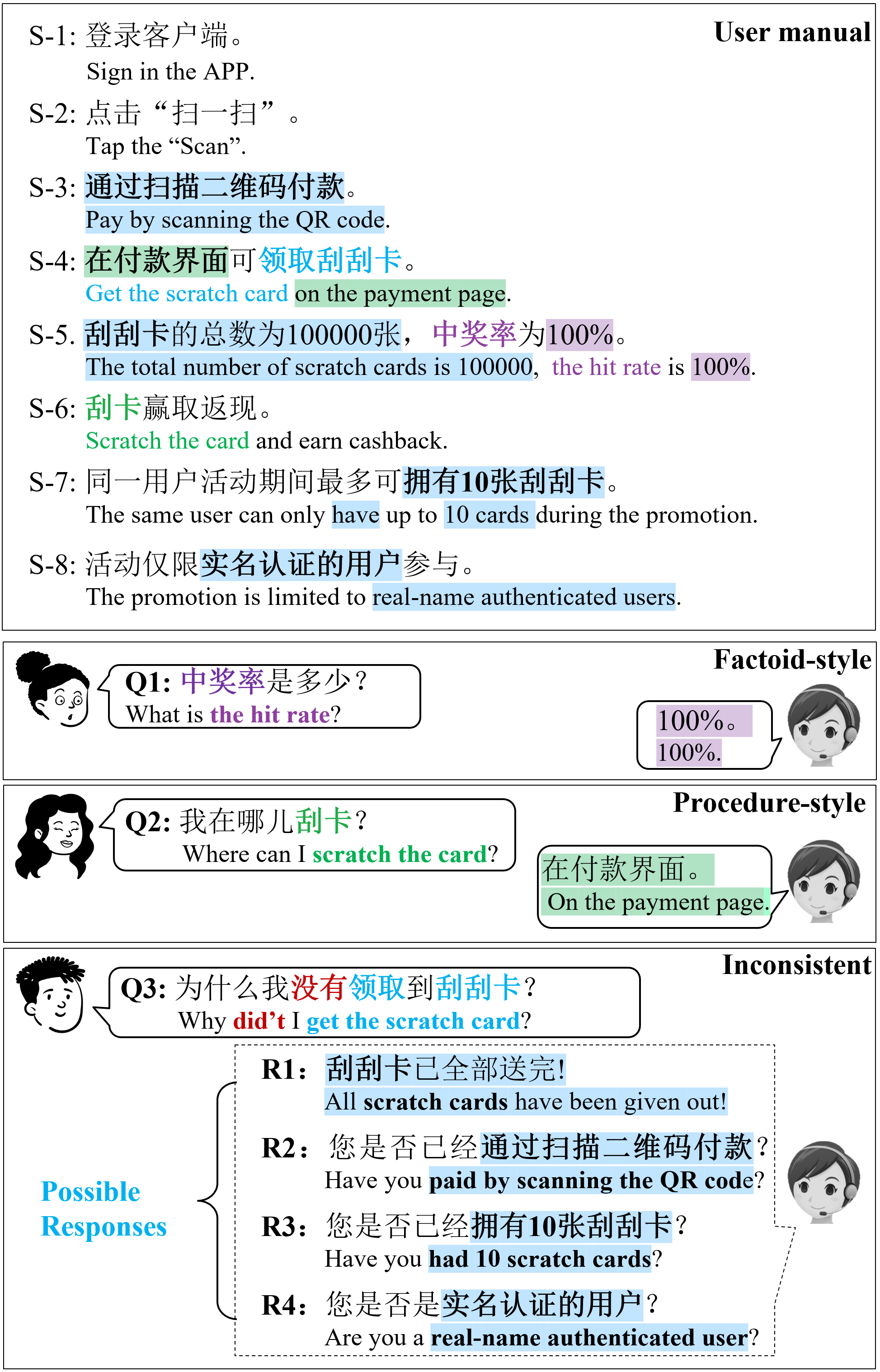}
    \caption{A user manual and three QA cases about it
    }
    \label{fig:style}
    \vspace*{-0.2cm}
\end{figure}
Let's see how a CSR approaches various questions given the user manual\footnote{We make a few modifications~(e.g., removing the application name) to this real-world user manual for anonymity.} in Figure~\ref{fig:style}. Specifically, she can answer Q1 by looking up the arguments of facts, i.e., the value (\colorbox{Purple!30}{100\%}) of \textcolor{Purple}{the hit rate}. For Q2, she needs to reason about the steps in the procedure as the answer is not explicitly exhibited in the user manual --- {the user manual directs the user to ``Sign in'' and ``Scan'', and eventually to the ``payment page''. After that, the subsequent description does not direct the user anywhere else. A user can infer from the procedure that one can scratch the card after immediately getting it on the payment page. 
Therefore, the location to \textcolor{Green}{scratch the card} is \colorbox{Green!30}{on the payment page}.} The most difficult one is Q3, which raises a case inconsistent with the descriptions in the user manual --- going along with the user manual~(S-4), the user \textcolor{Red}{didn't} \textcolor{Cerulean}{get the scratch card}. We call this type of question an {inconsistent question}. A possible response to the inconsistent question may be an answer~(R1) or a high-utility question leading to an answer~(R2-R4). Towards such responses, the CSR needs to reason about not only the steps but the facts binding the steps --- the inconsistency may be caused by incorrect operations in the previous step~(e.g., the user didn't \colorbox{Cerulean!30}{pay by scanning the QR code}) or unsatisfiable constraints in the facts~(e.g., the user already \colorbox{Cerulean!30}{had 10 scratch cards}). Therefore, aligned with the CSR, a practical MRC model must also be able to draw a unified inference across steps and facts to properly answer various user questions.
\par

With a splendid library of studies~\cite{rajpurkar2016squad,zhu2020question}, current MRC models have shown strong power to answer factoid-style questions~(e.g., Q1). 
More recent studies~\cite{tandon-etal-2020-dataset,goyal2021tracking,zhang2021knowledge} have mitigated the weakness of factoid-style MRC on procedure-style questions. However, these studies mainly focus on questions about the states of entities that can be answered by single spans. They have trouble answering more complex procedure-style questions with multiple-span answers or about the dynamics of actions~(e.g., Q2), let alone inconsistent questions like Q3. We also notice that existing models are only designed for single types of questions. This doesn't meet the real-world scenarios where factoid-style, procedure-style, and inconsistent questions are all involved~(cf., Table~\ref{tab:q_survey}).\par

\begin{figure*}[ht]
    \centering
    \includegraphics[width=\textwidth]{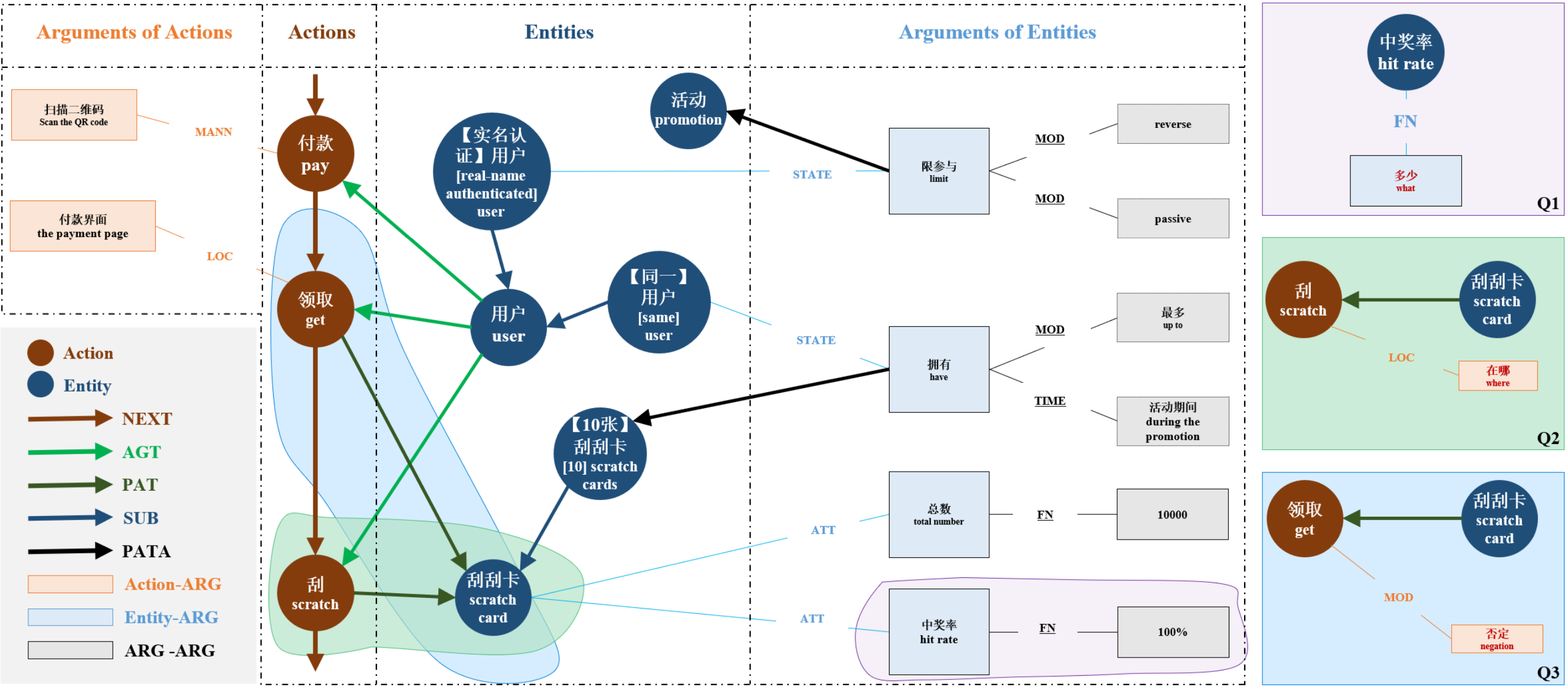}
    \caption{Screenshot of the graph of the user manual and the graphs of Q1, Q2, and Q3. The representation of a question and its most similar sub-graph are bounded in the same color and the conflict arguments are marked in red.}
    \label{fig:rep}
     \vspace*{-0.5cm}
\end{figure*}

To address these limitations, we propose the \textbf{\kh \& \kt} task\footnote{We name it after Gilbert Ryle's thought that the possession of \kh~(i.e., procedural knowledge) and \kt~(i.e., factoid knowledge) is ``a mark of intelligence''~\cite{ryle2009concept}.} --- given a user manual, the model is required to answer factoid-style, procedure-style or inconsistent questions about it. 
We resolve this task by jointly representing the s\underline{T}e\underline{P}s and f\underline{A}cts in a heterogeneous g\underline{RA}ph~(\rn), cf., Figure~\ref{fig:rep}. This representation allows us to make the unified inference of various questions. We further benchmark the proposed task with a \underline{H}euristic method~(HUM) to automatically represent \underline{U}ser \underline{M}anuals as \rns and a densely-annotated dataset~(\dataname) to test the model's ability in answering real-world questions. Specifically, HUM is designed to be independent of labelled data with the ultimate goal of building industry-scale applications and particular attention to low costs and good generalization capability. 
The annotated dataset~(\dataname) is derived from \underline{O}nline \underline{H}elp d\underline{O}cuments and real-world user questions from an e-commerce company. We provide the gold answers to user questions and gold \rns of user manuals.
Experiments demonstrate the superiority of \rn, the efficiency of HUM, and the significant challenges of \dataname. With an in-depth investigation of \rn, we discuss the issues and broader impact. We expect our work can advance the MRC of user manuals and inspire future research on smart customer service. We highlight our work as follows:
\begin{itemize}
    \item We introduce a new task and take the primary step to the MRC of user manuals in a more realistic setting, where various questions are involved.
    \item We resolve the task by jointly representing steps and facts in a graph. We also benchmark the task with an efficient method and densely annotate a testing set derived from real-world scenarios. 
    \item The experiment reveals the superiority of the proposed representation and the significant challenges of the new task. 
\end{itemize}
\begin{table*}[t]
    \centering
    \scalebox{0.77}{
    \begin{tabular}{l|l|c|c|c}
    \hline
    \textbf{Dataset} & \textbf{Types of user manuals} & \textbf{Number of user manuals} & \textbf{Number of annotations} & \textbf{QA pairs} \\
    \hline
        \citet{zhang2012automatically} & recipes, technical manuals & 74 & 1,979 & \textcolor{red}{\ding{55}} \\
        \hline
        \citet{mori2014flow} & recipes & 266 & 19,939 & \textcolor{red}{\ding{55}} \\
        \hline
        \citet{kiddon_mise_2015} & recipes & 133 & Not Avaliable & \textcolor{red}{\ding{55}} \\
        \hline
        \citet{yamakata2020english} & recipes & 300 & Not Avaliable & \textcolor{red}{\ding{55}} \\
        \hline
        \citet{jiang2020recipe} & recipes & 260 & 15,203 & \textcolor{red}{\ding{55}} \\
    \hline 
    \citet{nabizadeh2020myfixit} & technical manuals
 & 1,497 & Not Avaliable & \textcolor{red}{\ding{55}} \\
    \hline 
    \citet{goyal2021tracking} & technical manuals
 & 1,351 & 6,350 & \textcolor{red}{\ding{55}} \\
    \hline 
    \citet{zhong2020deep} & technical manuals
 & 400~(sentences) & 2,400 & \textcolor{red}{\ding{55}} \\
    \hline 
    \citet{mysore2019materials} & scientific experiment
 & 230 &19,281
 & \textcolor{red}{\ding{55}} \\
 \hline
 \citet{vaucher2020automated}&scientific experiment&1,764&$\sim$4,755&\textcolor{red}{\ding{55}}	\\
\hline			
\citet{kuniyoshi2020annotating}&scientific experiment&243&23,082&\textcolor{red}{\ding{55}}	\\
\hline			
\textbf{\dataname}&e-commercial helping documents&2,000&24,474	&\textcolor{Green}{\ding{51}}		\\

    \hline 
    \end{tabular}
    }
    \caption{Comparisons between \dataname with off-the-shelf datasets based on the statistics reported in the original paper}
    \label{tab:rel}
    \vspace*{-0.3cm}
\end{table*}

\section{Related Work}

\paragraph{MRC for user manuals} 
Earlier studies on the MRC of user manuals focus on factoid-style comprehension~\cite{zhang2012automatically,mysore2019materials,jiang2020recipe,nabizadeh2020myfixit}.
Recent research pays attention to the procedure-style comprehension of user manuals and aims to track the state values of pre-defined entities~\cite{bosselut2017simulating,amini2020procedural,tandon-etal-2020-dataset,goyal2021tracking,zhang2021knowledge}.
Although such models may answer some procedure-style questions, they are in trouble answering more complex questions like inconsistent ones. Several studies design structured representations for user manuals~\cite{kiddon_mise_2015,maeta2015framework,vaucher2020automated,kuniyoshi2020annotating}. However, they only represent the steps in user manuals. This largely limits the model's ability to answer questions that need inference about both steps and facts.
Contrary to the above research, we study the MRC of user manuals that are suitable for factoid-style, procedure-style, and inconsistent questions and design a heterogeneous graph to represent both steps and facts. The representation supports the unified inference of various questions and can be constructed automatically by a heuristic method.
\begin{table*}[t]
    \centering
    \scalebox{0.77}{
        \begin{tabular}{l|c|m{16cm}}
    \hline
      \textbf{Term} &\textbf{Category} & \textbf{Explanation} \\
      \hline
     Action & node & the action of a step performed by the user, presented as a verb, e.g., ``pay''.\\
        \hline
    Entity & node & the concepts in the user manual, e.g., ``scratch card''. Each user manual has a user Entity by default.\\
    \hline
 Action-ARG& argument& the modifier~(\textcolor{orange}{MOD}), time~(\textcolor{orange}{TIME}), location~(\textcolor{orange}{LOC}), and manner~(\textcolor{orange}{MANN}) arguments associated with an Action node. \\
    \hline
    Entity-ARG& argument& the arguments associated with an Entity node --- footnote argument~(\textcolor{NavyBlue}{FN}), offering extra details; attribute argument~(\textcolor{NavyBlue}{ATT}), describing a specific aspect~(e.g., the hit rate of the scratch card); state argument~(\textcolor{NavyBlue}{STATE}), describing a changeable state~(e.g., the ``have'' state of the same user).\\
    \hline
    ARG-ARG& argument& the arguments associated with an Entity-ARG. The arguments of \textcolor{NavyBlue}{ATT} are the same as Entity-ARG and the arguments of \textcolor{NavyBlue}{STATE} is the same as Action-ARG.\\
        \hline
    
    NEXT & relation & the directed edge between two Action nodes, indicating the end action is the next step of the start one. It determines the order of the actions performed by the user.\\
    \hline
    AGT&relation& the directed edge from the user Entity to an Action, indicating the action is performed by the user.
    \\
    \hline
    PAT& relation& the directed edge from an Action to an Entity, indicating the action is performed on the entity.\\
    \hline
    SUB& relation& the directed edge between two Entity nodes, indicating the start entity is a sub-entity of the end one.\\
    \hline
    PATA & relation& the directed edge from a \textcolor{NavyBlue}{STATE} to an Entity, indicating the entity is affected by the changing of \textcolor{NavyBlue}{STATE}.\\
    \hline
    \end{tabular}
    }
    \caption{Explanations about the nodes, relations, and arguments of \rn}
    \label{tab:kg}
    \vspace*{-0.5cm}
\end{table*}

\paragraph{Datasets for user manuals}
We have done a thorough survey~(cf., Table~\ref{tab:rel}) about existing datasets for user manuals. The types of user manuals include recipes~\cite{zhang2012automatically,mori2014flow,kiddon_mise_2015,yamakata2020english,jiang2020recipe}, technical manuals~(e.g., device  maintenance, surgical practices)~\cite{zhang2012automatically,nabizadeh2020myfixit,goyal2021tracking,zhong2020deep}, and scientific experiment~\cite{mysore2019materials,vaucher2020automated,kuniyoshi2020annotating}. There is a lack of a dataset about e-commerce scenarios where the MRC of user manuals is urgently needed due to the high labor costs for customer service. Besides, we have noticed that existing annotations are all about user manuals. None of them has offered real-world QA pairs to test the practical performance of MRC models. To address these limitations, we collect the \dataname dataset with online helping documents from an e-commerce company. We then contribute annotations for data-driven evaluations of the model's ability to answer real-world questions and represent user manuals.

\section{Representation of user manuals}
\label{sec:rep}

\paragraph{Definitions}
\label{sec:def} We wish to design a representation for user manuals that supports the unified inference of various questions. It can be reduced to two fundamental tasks: {how to extract actions~(entities) and their arguments from the user manual}, and {what is the relation between an action and an entity~(two actions or two entities)}. This motivates us to propose the \rn representation, which jointly describes steps and facts in a heterogeneous graph with two sets of nodes, three sets of arguments, and five sets of relations, as defined in Table~\ref{tab:kg}. In this way, the user manual in Figure~\ref{fig:style} is represented as a graph\footnote{Although the \textcolor{NavyBlue}{STATE} of Entity shares the same arguments with an Action node, it cannot be upgraded into a node because it doesn't refer to an action and cannot be linked to other Action nodes. Similarly, the \textcolor{NavyBlue}{ATT} of an Entity node cannot be a node because it can't be linked to other Entity nodes except for the parent one. Meanwhile, as the user dominates the whole user manual, it cannot be the target of any \textcolor{NavyBlue}{STATE}. }, part of which is displayed in Figure~\ref{fig:rep}. 

\paragraph{Unified inference}
\label{sec:unified}
We use \rn to represent Q1, Q2, and Q3, as shown in Figure~\ref{fig:rep}. The inference of answers can be divided into two stages. First, we extract a sub-graph from the representation of the user manual that is most similar to the structure of the question. Second, we identify the answers by directly inferring the conflict arguments between the sub-graph and the question\footnote{This operation reduces the cognition loads to identify Wh-word~(e.g., when, what) and its variants in user questions.}. 
Specifically, if the conflict is caused by the values of arguments, the answer is the value from the sub-graph. For example, the answer for Q1 is the \underline{FN} value~(``100\%'') of the hit rate. If the conflict is caused by the existence of Action-ARG, the answer is inferred from the nearest Action node, the same argument of which is not null, that can walk to the current Action node via NEXT relations. For example, the answer to Q2 is located in the \textcolor{orange}{LOC} argument of the ``get'' node, which is linked to ``scratch'' via NEXT. The third type of conflict is the inconsistency of \textcolor{orange}{MOD} arguments, namely, the Action node of the question has a \textcolor{orange}{MOD} argument valued ``negation''. The answer is inferred from the nearest Action node and Entity nodes pointed to the sub-graph. For example, all possible responses to Q3 are obtained from the ``pay'' node, the ``user'' node, and the ``scratch card'' node. In summary, \rn allows the unified inference of factoid-style, procedure-style, and inconsistent questions by jointly representing steps and actions in the same graph. 

\section{The HUM Method}
Towards practical industry-scale applications, it requires extra attention to labor costs and generalization. Thus, we choose to design an efficient method based on pre-trained semantic dependency parsing tools and domain-independent heuristics.\par
\begin{figure}[t]
    \centering
    \includegraphics[width=7cm]{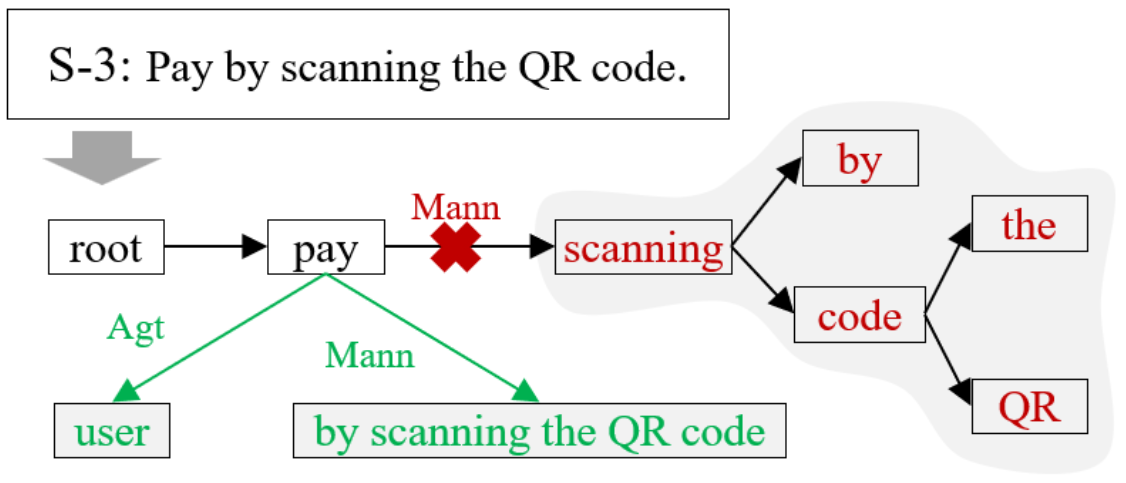}
    \caption{Modifications for the semantic dependency tree parsed from S-3. Red and green vertices are the eliminated and newly added vertices, respectively.}
    \label{fig:off}
\end{figure}
First, we segment the user manual sentence by sentence and feed the sentences into a pre-trained semantic dependence parsing tool. If a sentence is imperative, we leverage ``user'' as the subject of the sentence. For instance, ``Sign in the APP'' is modified as ``User sign in the APP''. Then, we eliminate the vertices to compose more meaningful phrases --- the offsprings of a predicate's child are eliminated and combined with the child. An example\footnote{To save space, some examples are only in one language.} is shown in Figure~\ref{fig:off}. In addition, to maintain the leading role of the user, if the user is the patient of a predicate, we change its role with the agent and mark the predicate as ``reverse''.  \par
    Second, we use the active verbs, whose agents are the user, as the Action nodes. These nodes are linked with the ``NEXT'' relations based on their order in the user manual and semantic dependencies. Besides the user node created by default, other Entity nodes are initialized as the unique patients with the user as agents\footnote{As the ``scratch card'' and ``card'' share the same semantics, we treat both of them as the ``scratch card'' entity. In addition, although ``limit'' is not an active verb, its patient~(``promotion'') is treated as an Entity node as its agent is the user.}. The relation from the user Node to any Action node is ``AGT''. We also add an edge with the ``PAT'' relation from an Action node to the Entity node, whose value serves as the patient of the Action node. Particularly, if an entity has attributives, we create a new entity with the ``SUB'' relation to it by combining the attributives. The ``same user'' node is created in this way. Most of Action-ARG, Entity-ARG, and ARG-ARG are generated by the alignment with the tags from off-the-shelf tools like LTP~\cite{che-etal-2021-n}, HanLP~\cite{he2020establishing}, etc. The alignment rules are defined in Table~\ref{tab:align}. Besides, if an entity acts as an attribute to an agent, we use the agent as an \textcolor{NavyBlue}{ATT} argument to the entity. For example, ``total number'' is an \textcolor{NavyBlue}{ATT} argument to the ``scratch card'' entity. We also use a predicate as the \textcolor{NavyBlue}{STATE} argument of its agent when the agent is the user, but the predicate is a state verb~( e.g., ``have'') or when the agent is not the user.
\begin{table}[t]
    \centering
    \scalebox{0.77}{
    \begin{tabular}{l|m{7.7cm}}
    \hline
        \textbf{ARGs} &  \textbf{Tags}\\
        \hline
        \textcolor{orange}{MOD} & mDEPD, mTime, mRang, mDegr, mFreq, mDir, mNEG, mMod\\
        \hline
        \textcolor{orange}{TIME} &Time, Tini, Tfin, Tdur, Trang\\
        \hline
        \textcolor{orange}{LOC} & Loc, Lini, Lfin, Lthru, Dir\\
        \hline
        \textcolor{orange}{MANN} & Mann, Tool, Matl, Accd \\
        \hline
        \textcolor{NavyBlue}{FN} & LINK, Clas\\
        \hline
    \end{tabular}
    }
    \caption{Alignment rules between the arguments and the tags from the semantic dependency parsing tools}
    \label{tab:align}
\end{table}

\paragraph{Automated Inference for QA} To infer the answers to user questions, we leverage the HUM method to represent the user manual and user question, simultaneously. Then, we extract the sub-graph that is most similar to the representation of the user question via the sub-graph matching algorithm~\cite{zou2011gstore}. The extraction only considers the nodes in the graph. The similarity score between two nodes~($x$ and $y$) is computed by 
\begin{equation}
\label{eq:sim}
\vspace*{-0.1cm}
    \mathcal{S}(x, y)=1-\mathcal{D}(x, y)/\max(|x|,|y|),
\vspace*{-0.1cm}
\end{equation}
where $\mathcal{D}$ computes the Levenshtein distance, and $|*|$ is the length of $*$. After getting the sub-graph matched to the user question, we do the inference described in Section~\ref{sec:rep} to get the answer.
\begin{figure}
    \centering
    \includegraphics[width=7cm]{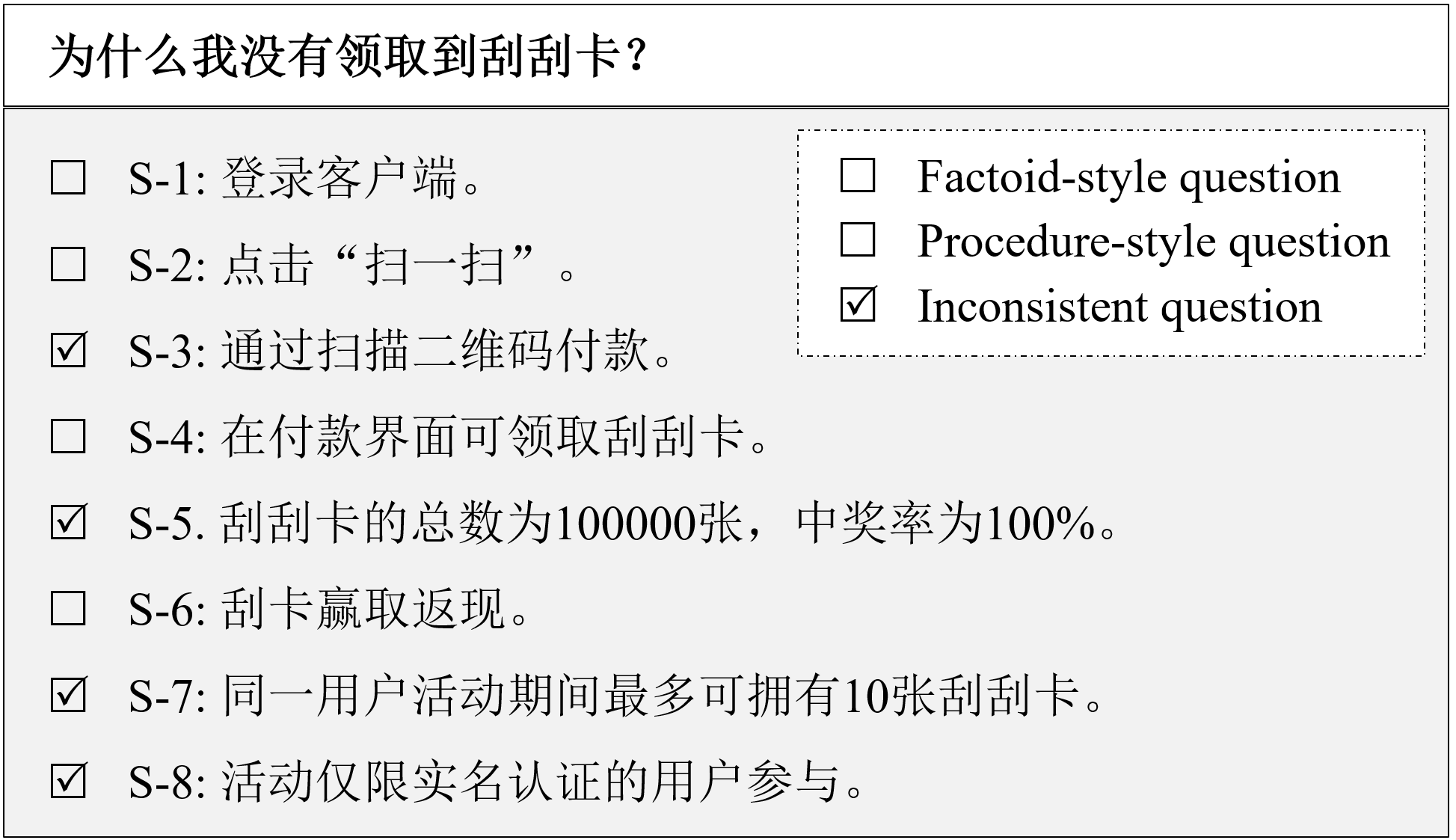}
    \caption{Annotation of the answer to Q3}
    \label{fig:ann}
\end{figure}
\begin{table}[t]
    \centering
    \scalebox{0.77}{
    \begin{tabular}{l|c|m{2cm}<{\centering}|m{1.8cm}<{\centering}}
    \hline
       \textbf{Type}  & \textbf{Number} & \textbf{Question Words}& \textbf{Answer spans} \\
       \hline
        Factoid-style & 955~(47.75\%) & 11.86& 1.17\\
        \hline
        Procedure-style & 483~(24.35\%) & 12.07& 1.14\\
        \hline
        Inconsistent & 562~(28.10\%) & 11.36& 1.16\\
        \hline
        All & 2000 & 11.77 & 1.16 \\
        \hline
    \end{tabular}
    }
    \caption{The number~(percentage) of questions, the average number of words in the question, and the average number of text spans in the answer}
    \label{tab:q_survey}
\end{table}


\section{The \dataname dataset}
We gather the frequently asked questions~(FAQs) summarized by the customer service department of an e-commerce company. We remove FAQs that aren't attached to any documents and that are attached to documents with only images. Finally, we get a dataset of 2000 user manuals, each of which has an FAQ.

\paragraph{Answer annotation} Each annotator is shown an FAQ and its attached user manual, which is displayed sentence by sentence. A toy example for the annotation of Q3 is shown in Figure~\ref{fig:ann}. The annotator is required to select which sentences in the user manual can be possible responses to the question and then select the type of the question. Each sentence is treated as a text span in the answer. The statistics of the annotations are described in Table~\ref{tab:q_survey}. It reveals that the procedure-style and inconsistent questions take up more than half part~(52.45\%) of the real-world user questions about user manuals.
\begin{table}[t]
    \centering
    \scalebox{0.77}{
    \begin{tabular}{|m{9.5cm}|}
    \hline
         B1: What are the actions that should be performed by the user in the manual?~(Action node) \\
         B2: What are the entities in the user manual?~(Entity node) \\
         B3: What are arguments~(modifier, time, location and manner) of the action?~(Action-ARG)\\
         B4: What are arguments~(footnote, attribute and state) of the entity?~(Entity-ARG)\\
         B5: What are the arguments of the attribute/state?~(ARG-ARG)\\
         B6: Is Action-2 in the next step of Action-2?~(NEXT)\\
         B7: Does the entity act as the patient of the action?~(PAT)\\
         B8: Is Entity-2 a sub-entity of Entity-1?~(SUB)\\
         B9: Does the entity act as the patient of the state~(PATA)\\
    \hline
    \end{tabular}
    }
    \caption{Basic questions for representation annotation}
    \label{tab:basic}
\end{table}

\paragraph{Representation annotation} This annotation is done on 200 user manuals of \dataname. As some user manuals describe more than one task,
there are 346 graphs that need to be annotated. It is not practical for the annotator to draw a heterogeneous graph for each user manual. Following \citet{dalvi2018tracking}, we reduce the task as the annotation of answers to nine basic questions derived from Table~\ref{tab:kg}. The basic questions are defined in Table~\ref{tab:basic}. Specifically, the annotations of B1 and B2 are completed first. This is because the annotations of B3-B4 and B6-B8 depend on the results of B1-B2.
Similarly, the annotations of B5 and B9 are completed last as they depend on B4. The statistics are shown in Table~\ref{tab:statistic}. Although the basic questions are factoid-style questions, we can use them and gold answers to roughly estimate the maximum potential of models --- if a model can correctly answer these basic questions, it can construct \rn for user manuals and thus have a chance to answer higher-level~(procedure-style, inconsistent or more) questions via unified inferences.
\begin{table}[t]
    \centering
    \scalebox{0.77}{
    \begin{tabular}{l|c|c|c}
    \hline
          &  Average & Minimum & Maximum\\
    \hline
    Sentences  /  user manual&7.73&2&33\\
    \hline 
    Words  /  sentence&40.91&10&205\\
    \hline
    Graphs  /  user manual & 1.73& 1 & 8\\
    \hline
    Actions  /  graph & 3.47&2&14\\
    \hline
    Entities nodes  /  graph & 2.95& 1 &6 \\
    \hline
    {Action-ARG}  /  Action &1.13&0&3  \\
    \hline
    {Entity-ARG}  /  Entity &1.40 &0 &3  \\
    \hline
    {ARG-ARG}  /  graph &0.86&0&3 \\
    \hline
    {NEXT}  /  graph &2.47 &1 &13 \\
    \hline
    {PAT}  /  graph &0.80 &0 &2 \\
    \hline
    {SUB}  /  graph &2.22 &0 &8 \\
    \hline
    {PATA}  /  graph &0.63 &0 &2 \\
    \hline
    \end{tabular}
    }
    \caption{Statistics of the representation annotation}
    \label{tab:statistic}
\end{table}
\paragraph{Quality control} We employ three CSRs with at least one-year working experience from the same department that summarizes the FAQs. We give a training session to the workers to help them fully understand our annotation requirements. The annotation tasks are released to the workers via an online annotation platform managed by the e-commerce company. On average, it takes about 3 minutes for an annotator to annotate the answer to an FAQ and about 2.8 hours to annotate all answers to basic questions of a user manual. An annotated result is accepted only if all three workers agree; otherwise, we invite two experts~(one is the manager of the customer service department and the other is a postdoc working on computational linguistics) to make the final decision. The experts closely discuss with each other to ensure the consistency of results.

\section{Experiments}
To systematically benchmark the \kh \&\kt task evaluation, we perform two experiments based on the \dataname dataset. First, with the help of answer annotations, we compare the performances of HUM with seven baselines in answering real-world user questions, i.e., the FAQ-answering task. Second, with the help of representation annotations, we investigate the maximum potential of HUM, i.e., the B-answering task. Specifically, we use the LTP tool~\cite{che-etal-2021-n} for word segmentation and semantic dependency parsing.

\subsection{Experiment I: FAQ-answering Task}
\paragraph{Baselines} We compare HUM with seven baselines: 1) QA pattern matching~\cite{peng2010formalized,jain2014rule}, which utilizes a number of QA patterns written by experts; 2) lexical matching~\cite{alfonseca2001prototype,yang2018adaptations}, which computes the lexical similarity scores between the FAQ and candidate sentence via EQ~(\ref{eq:sim}); 3) semantic matching, which computes similarity scores of user questions and candidate sentences based on BERT embeddings~\cite{devlin2018bert}; 4) keyword matching~\cite{moldovan2000structure}, which measures the similarity between the keywords of the user question and candidate sentence; 5) pre-trained LM~(i.e., PERT)~\cite{cui2022pert}, which has been trained with MRC tasks on other corpora; 6) self-supervised LM~\cite{nie2022unsupervised}, which leverages the self-supervised strategy to train the LM with more than 360000 user manuals\footnote{They are online helping documents from the e-commerce company without FAQs and annotations.} with interrogative masks~\cite{lewis2019unsupervised}; 7) fine-tuned LM, which fine-tunes the PERT models using a subset of \dataname, the results are obtained from two-fold cross-validation. We also create a variant of HUM with oracle \rns, named HUM-oracle. This method is tested on the 200 FAQs with user manuals that have gold \rns. 
\begin{figure*}
    \centering
    \includegraphics[width=\textwidth]{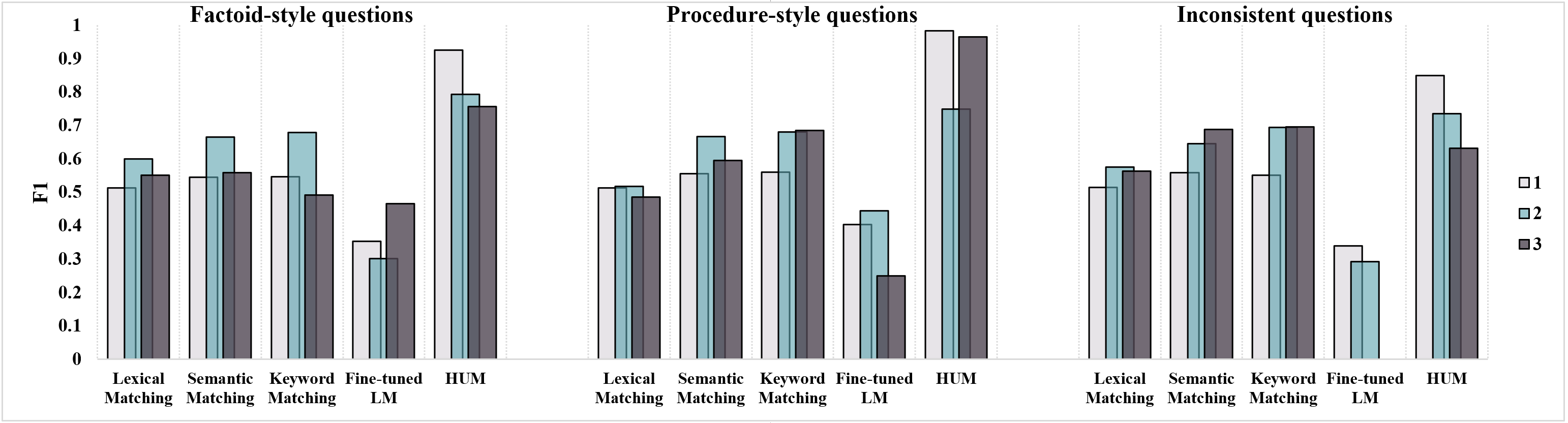}
    \caption{F1 scores w.r.t. different types of user questions and different number of answer spans}
    \label{fig:f1}
\end{figure*}

\paragraph{Evaluation metrics} We use the precision~(P), recall~(R), and F1 scores to measure the performances. To make fair comparisons with methods designed for one-span answers~(i.e., QA pattern matching, pre-trained LM, and self-supervised 
 LM), we also report P@1, R@1, and F1@1 values. 
\begin{table}[t]
    \centering
    \scalebox{0.77}{
    \begin{tabular}{l|c|c|c}
    \hline
     \textbf{Method}    &  P / P@1&R / R@1&F1 / F1@1\\
     \hline
      \textbf{QA pattern matching}   & 0.77&\textcolor{red}{0.18} &0.30\\
      \hline
      \textbf{Lexical matching}   & 0.45 / 0.50&0.56 / 0.46&0.50 / 0.48\\
      \hline
      \textbf{Semantic matching}& 0.36 / 0.38&0.60 / 0.36&0.45 / 0.37\\
      \hline
      \textbf{Keyword matching}   & 0.45 / 0.54&{0.74} / 0.41&0.56 / 0.52\\
      \hline
      \textbf{Pre-trained LM}  & \textcolor{red}{0.24} & 0.24&\textcolor{red}{0.24} \\
      \hline
      \textbf{Self-supervised LM} & 0.41 &0.34 &0.37 \\
      \hline
      \textbf{Fine-tuned LM}    & 0.57 / 0.64&0.26 / 0.23&0.36 / 0.34\\
      \hline
      \textbf{HUM} &{0.89} / {0.87}& 0.67 / {0.61}&{0.76} / {0.72}\\
      \hline
      \textbf{HUM-oracle} &\textcolor{Green}{0.97} / \textcolor{Green}{0.98}& \textcolor{Green}{0.93} / \textcolor{Green}{0.73}&\textcolor{Green}{0.95} / \textcolor{Green}{0.83}\\
      \hline
    \end{tabular}
    }
    \caption{Results of the FAQ-answering task. The values indicating the best performance and the worst performance are colored in green and red, respectively.}
    \label{tab:faq}
\end{table}

\begin{table*}[t]
    \centering
    \scalebox{0.8}{
    \begin{tabular}{c|m{3cm}<{\centering}|m{3cm}<{\centering}|m{3cm}<{\centering}|m{3cm}<{\centering}|m{3cm}<{\centering}}
    \hline
   \textbf{Basic Question}  & \textbf{Biaffine}&\textbf{MFVI}&\textbf{HanLP}&\textbf{LTP} & \textbf{Human}\\
     \hline
    \textbf{B1}& 0.23 / 0.40 / 0.29 & 0.27 / 0.53 / 0.36 &0.31 / 0.57 / 0.40  &0.34 / 0.70 / 0.46 &1.00 / 0.98 / 0.99\\
   \hline
   \textbf{B2}&0.10 / 0.08 / 0.09 & 0.20 / 0.16 / 0.18&0.17 / 0.15 / 0.16&0.24 / 0.25 / 0.25 & 0.90 / 0.98 / 0.94\\
   \hline
   \textbf{B3}& 0.09 / 0.08 / 0.08 & 0.19 / 0.19 / 0.19&0.19 / 0.21 / 0.20& 0.24 / 0.41 / 0.30 & 0.95 / 0.89 / 0.92\\
   \hline
   \textbf{B4}& 0.04 / 0.02 / 0.03  & 0.08 / 0.09 / 0.09& 0.06 / 0.06 / 0.06&0.12 / 0.16 / 0.14 & 0.88 / 0.80 / 0.84\\
   \hline
   \textbf{B5}&0.07 / 0.02 / 0.03 & 0.12 / 0.03 / 0.05&0.08 / 0.03 / 0.05&0.18 / 0.14 / 0.16 & 0.85 / 0.82 / 0.83\\
   \hline
   \textbf{B6}& 0.05 / 0.08 / 0.06 & 0.13 / 0.21 / 0.16&0.14 / 0.22 / 0.17& 0.18 / 0.35 / 0.24 & 0.98 / 0.98 / 0.98 \\
   \hline
   \textbf{B7}&0.16 / 0.14 / 0.15  & 0.26 / 0.35 / 0.30 &0.22 / 0.28 / 0.25&0.29 / 0.39 / 0.33 & 0.95 / 0.88 / 0.91\\
   \hline
   \textbf{B8}& 0.04 / 0.06 / 0.05 & 0.09 / 0.11 / 0.10&0.06 / 0.08 / 0.07&0.08 / 0.13 / 0.10 & 0.95 / 0.88 / 0.91\\
   \hline
   \textbf{B9}&0.25 / 0.06 / 0.10 & 0.34 / 0.13 / 0.19&0.24 / 0.05 / 0.09 &0.28 / 0.17 / 0.21 & 0.79 / 0.70 / 0.74\\
   \hline
   \textbf{Average} & 0.11/ 0.10 / 0.10 & 0.19 / 0.20 / 0.18 & 0.16 / 0.18 / 0.16 & 0.22 / 0.30 / 0.24& 0.92 / 0.88 / 0.90 \\ 
   \hline
    \end{tabular}
    }
    \caption{Performance of HUM using different semantic parsing tools and human w.r.t. the basic questions. For B1-B5, we report the P$\dag$ / R$\dag$ / F1$\dag$ values. For B6-B9, we report the P / R / F1 values.}
    \label{tab:basic_a}
    \vspace*{-0.2cm}
\end{table*}
\paragraph{Results} Table~\ref{tab:faq} shows the performance of HUM and baselines on the FAQ-answering task. We can see that HUM-oracle significantly outperforms others on all metrics. This demonstrates that representing user manuals as \rns is a desired resolution to answer various user questions. We also make the following observations. 1)~The keyword matching method gets a higher recall score than HUM, but its precision score is about half HUM. The matching of keywords between the user question and the user manual is similar to the extraction of the sub-graph from the heterogeneous graph of the user manual. So, this method can find all related spans to the user question. However, without the unified inference on \rn, it cannot reject the noise spans and thus get low precision scores. Similar results can also be found in the other matching methods. 2)~The QA pattern matching has a high precision score because of sophisticated QA patterns~(covering  more than 80\% of FAQs) but the worst recall score due to the high flexibility of answer expressions. 3)~We are not surprised that the pre-trained LM and self-supervised LM rank last, as they are only exposed to factoid-style questions during the training phase thus they are insufficient to answer procedure-style and inconsistent questions.  4)~The results of fine-tuned LM are slightly better than that of pre-trained LM and self-supervised LM but are still far behind HUM. We conjecture this is because of the lack of massive annotated data and the lack of unified inference. 5)~After carefully checking HUM, we notice that about 5\% of FAQs are not answered because HUM can't extract sub-graphs from user manuals. This suggests a possible future work is to eliminate such errors via more sub-graph matching algorithms, etc. 6)~Except for HUM-oracle, the best F1 score is only 0.76, indicating the intrinsic challenge of the \kh \& \kt task and there is substantial room for a model to represent user manuals as \rns like an expert.

\paragraph{Auxiliary results} We conduct a further study about methods that can produce multiple-span answers. As shown in Figure~\ref{fig:f1}, in most cases, the performances decrease with the increase of answer spans. This suggests that it is more difficult to predict answers with more spans than answers with fewer spans. Compared with other methods, the fine-tuned LM rank last on all metrics and has poorer robustness when answering different types of user questions --- the performance for inconsistent questions is significantly worse than that for procedure-style and factoid-style questions. It indicates that, despite the achievement of state-of-the-art results on many NLP tasks, 
the popular fine-tuned LM is not suitable for industry-scale MRC of user manuals without spending huge costs to annotate massive data. Notably, for all types of user questions, the F1 scores of HUM surpass other methods by a large margin on one-span and two-span answers and are slightly lower than the best value on three-span answers. This reveals the sufficiency and robustness of HUM to cope with the challenges raised by various types of user questions and multiple-span answers. 

\subsection{Experiment II: B-answering Task}
We take a close look at HUM to investigate its ability to represent user manuals as \rns and its maximum potential to answer higher-level questions.
We create variants of HUM by replacing the backbone semantic parsing tool~(i.e., LTP) with Biaffine~\cite{dozat2018simpler}, MFVI~\cite{wang2019second}, and HanLP~\cite{he2020establishing}. In addition, we employ five postgraduate students who major in computer science as annotators. They are asked to annotate the answers to basic questions. Before starting
the annotation work, they are shown 10 samples annotated by experts without attending the training session. In this way, we obtain human~(non-expert) upper bounds for this task. 

\paragraph{Evaluation metrics} Inspired by the evaluation of information extraction, we employ the precision, recall, and F1 scores based on BLEU scores~(denoted as P$^\dag$, R$^\dag$, and F1$^\dag$, respectively)~\cite{tandon-etal-2020-dataset} to measure the performances of HUM on B1-B5. For B6-B9, we report the standard precision, recall, and F1 scores.
\paragraph{Results} The results are reported in Table~\ref{tab:basic_a}. The last column reports the human upper bounds. Although we have tried four state-of-the-art semantic parsing tools, the best-performing method only reaches $\sim$0.24 F1 scores, indicating the significant challenges to automatically constructing \rns for user manuals. Specifically, the results of B6 are lower than that of B1, including the human results. This is because a worker's annotations to B6 are based on his annotations to B1, namely, the accumulative errors issue. The values of the human upper bound~($\sim$0.92 precision score, $\sim$0.88 recall score and $\sim$0.90 F1 score) demonstrate that the task is feasible, well-defined and leaves plenty of room for more advanced semantic dependency parsing tools, more efficient heuristics, etc.
\paragraph{Issues and broader impact of \rn} After investigating the bad cases, we find that, in addition to the accumulative errors, there are issues with co-reference problems, complex discourse parsing beyond sentences, etc. We leave the resolutions to these issues as future work. We observe that the performances of HUM on the B-answering task are much worse than that on the FAQ-answering task. Although the basic questions are factoid-style questions, they can compose all possible higher-level questions besides procedure-style and inconsistent ones. For example, Q2 can be decomposed into a combination of B1, B3, and B6. The FAQs in the B-answering task only take a small part of the questions composed from the basic questions. Thus, the B-answering task is much more challenging than the FAQ-answering task. This observation also strongly demonstrates that better performance on the B-answering task will largely improve HUM's ability to answer real-world questions. Besides, we'd like to talk about other potential benefits that can be gained from the \rns. In addition to the unified inference of various questions, the joint representation of dynamic actions and entities also sheds light on the reasoning and planning of new tasks. Possible applications involve the arrangement of unordered actions, error detection of the draft user manuals, the automated composition of user questions, task-oriented information seeking, etc. The model, with the ability to correctly answer basic questions, can further be applied in downstream scenarios other than QA. For example, it is beneficial to build an intelligent training bot for new staff from customer service departments.

\section{Conclusion and Future work}
We propose the \kh \& \kt task that requires the model to answer various questions about a user manual. We resolve it by representing the steps and facts of a user manual in a unified graph~(\rn). To benchmark it, we design an efficient method and annotate a testing set derived from real-world customer service scenarios. Experiments reveal the superiority of TARA, the efficiency of HUM, and the significant challenges of OHO and the new task.
\par
We take the primary step to study the MRC of user manuals in a more challenging setting, where various questions are involved. We hope our work can benefit further research on smart customer service. There are several directions for further work. First, to improve the performance of HUM, we will research resolutions to accumulative errors, co-reference problems, complex discourse parsing beyond sentences, etc. Second, to explore the potential of the proposed representation, we will introduce it to other tasks that need better interpretability, like task-oriented information seeking. Third, we will investigate the potential of unified inference for more complex user manuals, e.g., user manuals with multiple agents. Lastly, we plan to deploy HUM to online customer service to gain more insights for further improvements.

\clearpage
\section{Limitations}
We now explain the limitations and potential risks of our work. First, it seems the \kh \& \kt task is a bit unfriendly to supervised learning methods as we only annotate the testing set. However, towards practical  
industry-scale applications, we encourage future work to utilize the current annotations and contribute to more efficient heuristic, unsupervised, self-supervised, or weakly supervised methods, etc. Second, each user manual in \dataname only contains one user~(agent). However, there are a number of user manuals involving more than one agent, e.g., ``invite your friend as a new user and get cash back''. This motivates us to explore multi-agent user manuals in our future work. Third, in addition to the textual content, many user manuals contain visual information like images and GIFs. Hence, it will be more desirable to add such user manuals and study the \kh \& \kt task in multi-modal settings.


\section{Ethics Statement}
This paper presents a new task for machine comprehension of user manuals. Although the user manuals and FAQs involved in the task are collected from an e-commerce company, they are designed for normal users and have been widely used by the public for some time. We also have carefully checked these data to make sure they don't contain any personally identifiable information or sensitive personally identifiable information. Thus, we believe there are no privacy concerns.\par
All user manuals and FAQs are reviewed at least three times by the company's staff before being released to the public. Besides, we have been authorized by the company to make \dataname publicly available for academic research. Thus, we believe the dataset doesn't contain any harmful information and is qualified for distribution.\par
The annotators of \dataname consist of CSRs, a postdoc, and undergraduate students. As the dataset is about user manuals and the job is to answer questions about the user manuals, we believe there are no physical or mental risks to the annotators. 
\section*{Acknowledgement}
This work was supported in part by the National Natural Science Foundation of China (No. 62206191 and No. 62272330);
 in part by the China Postdoctoral Science Foundation (No.2021TQ0222 and No. 2021M700094); 
in part by the Natural Science Foundation of Sichuan (No. 2023NSFSC0473), 
and in part by the Fundamental Research Funds for the Central Universities (No. 2023SCU12089 and  No. YJ202219).

\bibliographystyle{acl_natbib}
\bibliography{anthology}
\clearpage
\appendix

\section{Settings of the baselines for the FAQ-answering task}

The detailed settings of the baselines models of the FAQ-answering task are as follows.
\begin{itemize}
    \item QA pattern matching~\cite{peng2010formalized,jain2014rule} uses handwritten rules to match questions and answers with corresponding syntactic formats~\cite{ravichandran2002learning, soubbotin2001patterns, greenwood2003using}. For example, given Q1 following the pattern of~\underline{what be <entity>?}, the answer is the sentence in the user manual following the pattern of {\underline{<entity> be <value>}}.
    
    \item Lexical matching~\cite{alfonseca2001prototype,yang2018adaptations}. We use EQ~(\ref{eq:sim}) to calculate the lexical similarity between question and candidate sentence~\cite{coates2022identifying}. Then we choose the top two sentences with the highest score as the final answer.
    
    \item Semantic matching~\cite{devlin2018bert}. We represent a sentence by averaging the token embeddings in this sentence. Then we compute the cosine similarity between question representation and candidate representation. The top two sentences with the highest score are used as the final answers.
    
    \item Keyword matching~\cite{moldovan2000structure}. We use the TF-IDF algorithm to extract ten keywords for each user manual. Meanwhile, we use the keywords extraction API of the iFLYTEK open platform to obtain the keywords of the question text. We calculate the matching score between each candidate answer and question according to the following formula:
    \begin{equation}
    \small
    \begin{aligned}
    \mathcal{S}=&16 \times |K_Q|+  16\times |K_A|+  16\times |K_Q \cap K_A|+ \\ 
    &16\times |\Gamma_{K_Q} \cap \Gamma_{K_A}|-  4\times \sqrt{\mathcal{D}_{\max({K_{\{Q\}}})}}
    \end{aligned}
    \end{equation}
    where $|K_Q|$ represents the number of candidate answer keywords existing in the question text, $|K_A|$ represents the number of question keywords existing in the candidate answers, $|K_Q \cap K_A|$ represents the number of words that are both question keywords and answer keywords, $\Gamma_{K_Q} \cap \Gamma_{K_A}$ represents how many question keywords are on the same sub\_tree of the candidate answer parsing tree, where the parsing tree of the candidate answer is obtained from the semantic dependency parsing of LTP~\cite{che-etal-2021-n}, $\mathcal{D}_{\max({K_{\{Q\}}})}$ represents the distance between the two question keywords that are farthest apart in the candidate answer text.
    
    \item Pre-trained LM. We use the Chinese machine reading comprehension model based on PERT-large~\cite{cui2022pert}, which has been fine-tuned on a mixture of Chinese MRC datasets. It is highly competitive in many tasks of machine reading comprehension and sequence labeling. It obtained 90.8, 95.7, and 79.3 F1-score on CMRC 2018 Dev, DRCD Dev, and SQUAD-Zen Dev (Answerable) data set respectively.
    We concatenate the question text and the user manual and then feed it to the model to predict the start and end positions of answer spans. 
    
    \item Self-supervised LM~\cite{nie2022unsupervised}. Self-supervised learning is the approach that trains the model using the constructed data sets of potential QA data mined from a large amount of  corpus~\cite{baralunsupervised}. It can obtain a quantity of data without manual annotation to fine-tune the pre-training model. Inspired by previous work~\cite{lewis2019unsupervised}, we mask actions, entities, and corresponding arguments and relations in the source user manuals and replace them with interrogative words (when, where, why, how, etc.) to construct QA data. We finally constructed 800,000 QA pairs from 360,000 user manuals as the training dataset. For model training, we still choose the PERT-large model, whose pre-training paradigm is also self-supervised learning. Following the pre-training paradigm, we concatenate the question text and the original user manual as the input of the model and then predict the start and end positions of the answer span. We choose the AdamW~\cite{loshchilov2017decoupled} as the optimizer, the learning rate is set to 1e-5, and a total of three epochs are trained. 
    
    \item Fine-tuned LM. The labelled dataset \dataname is divided into two parts. We fine-tune the pre-training model on half of the data and then evaluate the performance of the model on the other half of the data. To ensure that the model can output multiple possible answer spans for each question, we splice the question text and candidate sentences in the source user manual into the model in turn and then predict whether each candidate sentence is the answer to the question, and we still choose PERT-large as the pre-training model. The GPU we used for model training is Tesla P100, the max length of the model input is set to 512, the batch size is set to 4, the learning rate is set to 1e-5, and a total of six epochs are trained. The results are obtained from two-fold cross-validation.

\end{itemize}

\section{Case Study of the B-answering task}
We here discuss the issues of HUM in generating \rns for user manuals. Figure 6 presents a bad case caused by the co-reference problem --- the model fails to identify that ``they'' refers to ``real-name authentication users'', leading to a missing \textcolor{NavyBlue}{STATE} argument of the ``real-name authentication users'' node. Figure 7 presents a bad case caused by accumulative errors --- the model generates noise relations in \rn after it wrongly treats ``pass'' as an Action node.
\begin{figure}[h]
    \centering
    \includegraphics[width=7cm]{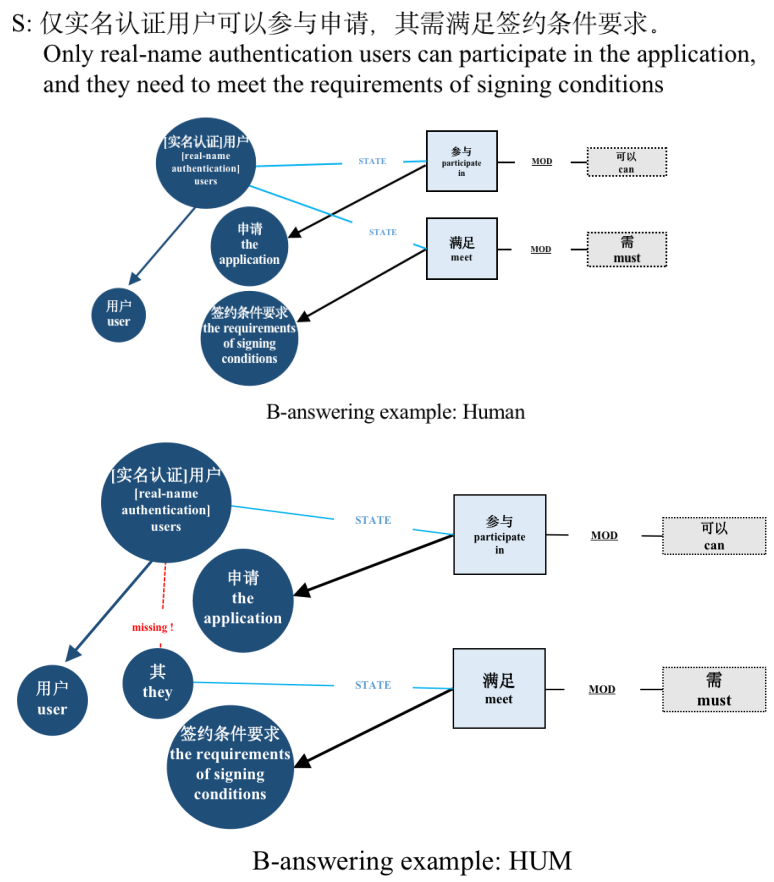}
    \caption{Bad case caused by the co-reference issue}
\end{figure}

\begin{figure}[h]
    \centering
    \includegraphics[width=6cm]{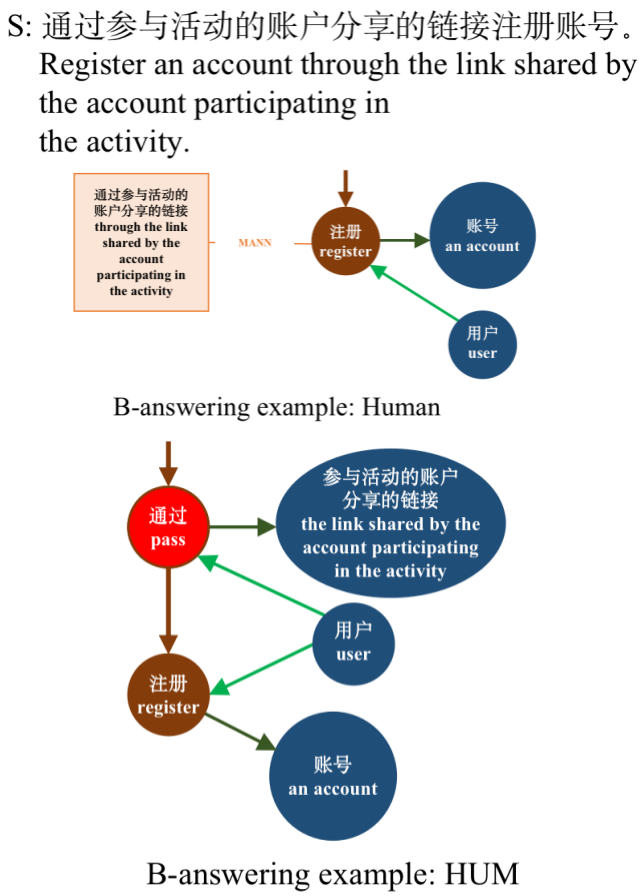}
    \caption{Bad case caused by accumulative errors}
\end{figure}

\end{document}